\documentclass{article}

\usepackage{arxiv}

\usepackage[utf8]{inputenc} 
\usepackage[T1]{fontenc}    
\usepackage[hidelinks]{hyperref}       
\usepackage{url}            
\usepackage{booktabs}       
\usepackage{amsfonts}       
\usepackage{nicefrac}       
\usepackage{microtype}      
\usepackage{lipsum}		
\usepackage{graphicx}
\usepackage[square,sort,comma,numbers,compress]{natbib}
\usepackage{doi}
\usepackage{amsmath}

\title{
\begin{center}
Subtle Data Crimes: 
\end{center}
Naively training machine learning algorithms could lead to overly-optimistic results}

\author{ \href{https://orcid.org/0000-0002-2267-0561}{\includegraphics[scale=0.06]{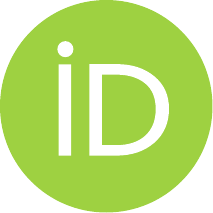}\hspace{1mm}Efrat Shimron}\thanks{Department of Electrical Engineering and Computer Sciences, UC Berkeley, Berkeley, CA, 94720 } \\
	UC Berkeley \\	
	\texttt{efrat.s@eecs.berkeley.edu} \\

	\And
	
	\href{https://orcid.org/0000-0001-9113-9566}{\includegraphics[scale=0.06]{orcid.pdf}\hspace{1mm}Jonathan I. Tamir}\thanks{Department of Electrical and Computer Engineering \& Department of Diagnostic Medicine, Dell Medical School \& Oden Institute for Computational Engineering and Sciences, The University of Texas at Austin, Austin, TX, 78712} \\
	UT Austin\\
	\texttt{jtamir@utexas.edu} \\
	
	\And
	
	\href{https://orcid.org/0000-0001-5951-1727}{\includegraphics[scale=0.06]{orcid.pdf}\hspace{1mm}Ke Wang}\footnotemark[1] \\
	UC Berkeley\\
	\texttt{kewang@berkeley.edu} \\

	\And
	
	\href{https://orcid.org/0000-0002-4794-221X}{\includegraphics[scale=0.06]{orcid.pdf}\hspace{1mm}Michael Lustig}\footnotemark[1] \\
	UC Berkeley\\
	\texttt{mlustig@eecs.berkeley.edu} \\
}

\renewcommand{\shorttitle}{\emph{Subtle Data Crimes}}

\hypersetup{
pdftitle={Subtle Data Crimes},
pdfsubject={Naively training machine learning algorithms could lead to overly-optimistic results},
pdfauthor={Efrat Shimron, Jonathan Tamir, Ke Wang, Michael Lustig},
pdfkeywords={Subtle Data Crimes, inverse crime, MRI, reconstruction, algorithms, bias},
}

\begin{document}
\maketitle

\begin{abstract}
	While open databases are an important resource in the Deep Learning (DL) era, they are sometimes used "off-label": data published for one task are used for training algorithms for a different one. This work aims to highlight that in some cases, this common practice may lead to \emph{biased, overly-optimistic} results. We demonstrate this phenomenon for inverse problem solvers and show how their biased performance stems from hidden data processing pipelines. We describe two data processing pipelines typical of open-access databases and study their effects on three well-established algorithms developed for Magnetic Resonance Imaging (MRI) reconstruction: Compressed Sensing (CS), Dictionary Learning (DictL), and DL. In this large-scale study we performed extensive computations. Our results demonstrate that the CS, DictL and DL algorithms yield systematically biased results when naively trained on seemingly-appropriate data: the  Normalized Root Mean Square Error (NRMSE) improves consistently with the data processing extent, showing an artificial increase of $25\%$-$48\%$ in some cases. Since this phenomenon is generally unknown, biased results are sometimes published as state-of-the-art; we refer to that as \emph{subtle data crimes}. This work hence raises a red flag regarding naive off-label usage of Big Data and reveals the vulnerability of modern inverse problem solvers to the resulting bias. 
\end{abstract}


\section{Introduction}
Public databases are an important driving force in the current Deep Learning (DL) revolution; ImageNet \cite{russakovsky2015imagenet} is a well-known example. However, due to the growing availability of open-access data and the general hype around AI, databases are sometimes used in an "off-label" manner: data published for one task are used for different ones. Here we aim to show that such na\"ive and seemingly-appropriate usage of open-access data could lead to biased, overly-optimistic results. 

Biased performance of machine learning models due to faulty construction of data cohorts or research pipelines has been recently identified for various tasks, including gender classification \cite{buolamwini2018gender}, COVID-19 prediction \cite{wynants2020prediction} and natural language processing \cite{mehrabi2021survey}. However, to the best of our knowledge, it was not yet studied for inverse problem solvers. We address this gap by highlighting scenarios that lead to biased performance of algorithms developed for image reconstruction from undersampled Magnetic Resonance Imaging (MRI) measurements; the latter is a real-world example of an inverse problem and a current frontier of DL research \cite{antun2020instabilities,zhu2018image,hammernik2018learning,wang2016accelerating,wang2018image, lundervold2019overview,mazurowski2019deep,knoll2020deep,ravishankar2019image}.

\begin{figure*}[!htb]
\centering
\includegraphics[width=1.05\textwidth]{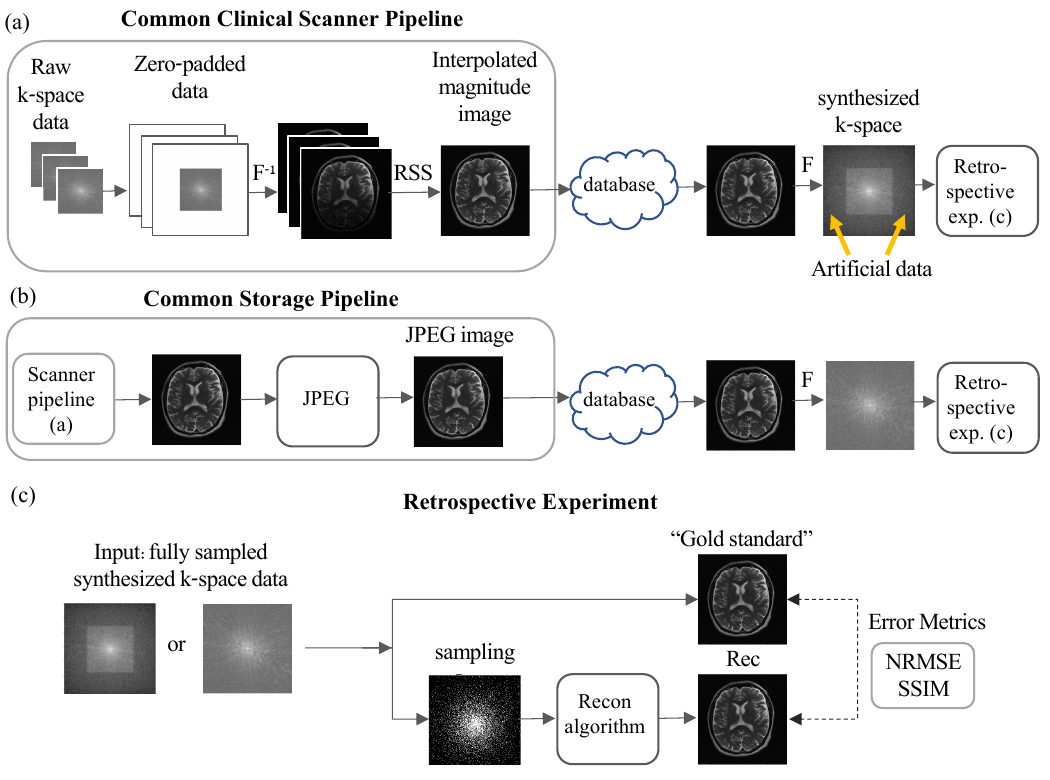}
\caption{\emph{Subtle data crimes}: how retrospective subsampling of processed data leads to biased results. (a) A common data processing pipeline, which is often implemented inside commercial MRI scanners, includes: k-space zero-padding, application of the inverse Fourier Transform, and coil combination via a Root Sum-of-Squares (RSS) step. The output image, which is interpolated and non-negative, is stored in a database. In \emph{subtle data crime I}, this image is later used for synthesizing new k-space data; this yields artificial data in previously zero-padded areas. (b) A common data storage pipeline includes JPEG compression. In \emph{subtle data crime II}, the compressed image is later used for retrospective experiments. (c) Standard research pipelines commonly involve retrospective subsampling of fully-sampled k-space data. In the \emph{subtle data crimes} scenarios, the fully-sampled data are based on processed data, hence image reconstruction algorithms benefit from the early data processing. Moreover, since the "gold standard" image is based on the same processed data as the reconstructed one, error metrics become blind to the data processing and they are therefore also prone to bias.}
\label{fig:fig1}
\end{figure*}

MRI measurements are fundamentally acquired in the Fourier domain, which is known as "k-space". Sub-Nyquist sampling is commonly applied for shortening the traditionally lengthy MRI scan time, and image reconstruction algorithms are used for recovering images from the undersampled data \cite{nishimura1996principles,mcgibney1993quantitative,griswold2002generalized,pruessmann1999sense}. The development of such algorithms should therefore ideally be done using raw k-space data. However, the development of DL methods requires thousands of examples, and databases containing raw k-space data are scarce. To date, there are only several databases that offer such data, e.g. \cite{knoll2020fastmri,ong2018mridata,calgary,desai2021skm}, while there are many more that offer reconstructed and processed Magnetic Resonance (MR) images, e.g. \cite{IXI,human_connectome@2021,AccelMR,oasis,ADNI,biobank,TCIA,brain_tumor_data}. The latter offer images for post-reconstruction tasks such as segmentation and biomarker discovery. Nevertheless, due to their abundance, they are often downloaded and used for synthesizing "raw" k-space data using the forward Fourier transform; the synthesized data are then used for the development of reconstruction algorithms. We identified that this common approach can lead to undesirable consequences; the underlying cause is that the non-raw MR images are commonly processed using hidden pipelines. These pipelines, which are implemented by commercial scanner software or during database storage, include a full set or a subset of the following steps: image reconstruction, filtering, storage of magnitude data only (i.e. loss of the MRI complex values), lossy compression, and conversion to DICOM or NIFTI formats; these reduce the data entropy. We aim to highlight that when modern algorithms are trained and evaluated using such data, they benefit from the data processing and hence tend to exhibit overly-optimistic results as compared to performance on raw, unprocessed data. Since this phenomenon is largely unknown, such biased results are sometimes published as state-of-the-art, without reporting the data processing pipelines or addressing their effects. In order to raise community awareness to this growing problem, we coin the term \emph{subtle data crimes} to describe such publications, in reference to the more obvious \emph{inverse crime} scenario \cite{colton1998inverse} which is described next.

Bias stemming from the underlying data has been previously recognized in a few scenarios related to inverse problems. The term \emph{inverse crime} describes a scenario in which an algorithm is tested using simulated data, and the simulation resonates with the algorithm such that it leads to improved results \cite{colton1998inverse,kaipio2007statistical,hansen2010discrete,guerquin2011realistic,mueller2012linear}. Specifically, the authors of \cite{guerquin2011realistic} described an \emph{inverse crime} as a situation where the same discrete model is used for simulating k-space measurements and reconstructing an MR image from them; they showed that this leads to reduced ringing artifacts compared with reconstruction from raw or analytically-computed measurements. A second example is evaluation of MRI reconstruction algorithms on real-valued magnitude images; in this case k-space exhibits conjugate symmetry, hence it is sufficient to use only about half of it for full image recovery. This symmetry is often leveraged in Partial Fourier methods such as Homodyne \cite{mcgibney1993quantitative} and POCS \cite{haacke1991fast}, where additional steps are applied for recovery of the full complex data. However, neglecting the data complexity creates a better-conditioned inverse problem and may hence lead to an obvious advantage when evaluating the algorithm on such data as opposed to raw k-space data. However, to the best of our knowledge, \emph{inverse crimes} were not yet studied in the context of machine learning or public data usage.

Here we introduce two subtle forms of algorithmic bias that were not yet considered and are relevant to the current DL era. We show how they arise from two hidden data processing pipelines that characterize many open-access MRI databases: a commercial scanner pipeline and a JPEG data storage pipeline. To demonstrate these scenarios, we took raw MRI data and "spoiled" them with carefully-controlled data processing steps; we then used the processed datasets for training and evaluation of algorithms from three well-established MRI reconstruction frameworks: Compressed Sensing (CS) with a Wavelet transform \cite{lustig2007sparse}, Dictionary Learning (DictL) \cite{ravishankar2010mr}, and DL \cite{aggarwal2018modl}. Our large-scale experiments demonstrate that these algorithms yield overly-optimistic results when trained and evaluated on processed data. Preliminary results of this work were published in a conference proceeding \cite{shimron2020subtle}.

The main contributions of this work are threefold. First, we reveal scenarios in which algorithmic bias of inverse problem solvers may arise from off-label usage of open-access databases. We also analyze the effects of \emph{inverse crimes} on complex high-dimensional learning systems via large-scale statistics. Secondly, we expose that CS, DictL and DL algorithms are all prone to this form of subtle bias. While recent studies identified stability issues of MRI reconstruction algorithms \cite{antun2020instabilities,darestani2021measuring}, to the best of our knowledge this is the first study that identifies a common vulnerability of canonical algorithms to such data-related bias. Third, by introducing the concept of \emph{subtle data crimes} and setting a framework for studying them, we hope to raise community awareness to the growing problem of bias stemming from off-label usage of open access data.

\begin{figure*}[ht]
\centering
\includegraphics[width=1.05\textwidth]{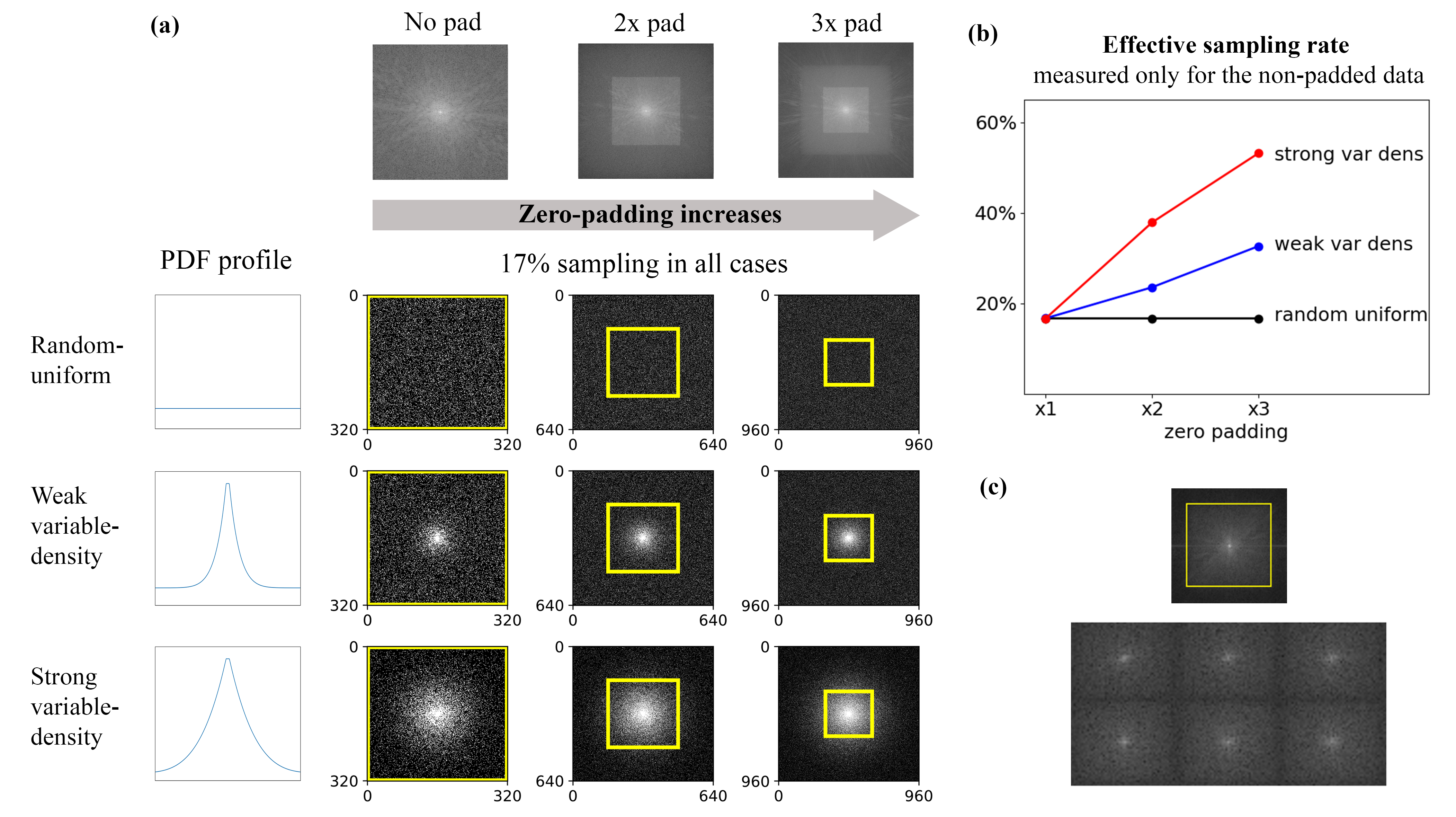}
\caption{An experiment demonstrating how retrospective-subsampling of k-space that was synthesized from processed data leads to increased effective sampling density of the "true" k-space data. (a) Subsampling masks were generated for different combinations of zero-padding factors (left-right) and subsampling schemes (top-down). The masks were generated from symmetric 2D Probability Density Functions (PDFs) (profiles displayed), with 17\% sampling in all cases. The regions covering of the original non-padded k-space data are marked with yellow boxes. Notice that the zero-padding squashes the original data to the center, so when a variable density scheme is used, those data are sampled with an increased rate. (b) Effective sampling rate, which is the subsampling rate inside the original k-space area (yellow boxes in (a)), vs. the zero-padding. Notice that for the VD schemes, the effective rate is much higher than the global rate (17\%) and may rise above 55\%. (c) Real-world examples for k-space data generated from MR images found in public open-access data \cite{IXI,human_connectome@2021} show evidence of zero-padding (the yellow box is our estimation). These examples indicate that training algorithms using data from public databases could lead to increased effective sampling.}
\label{fig:fig2}
\end{figure*}

\begin{figure*}[ht]
\centering
\includegraphics[width=1.05\textwidth]{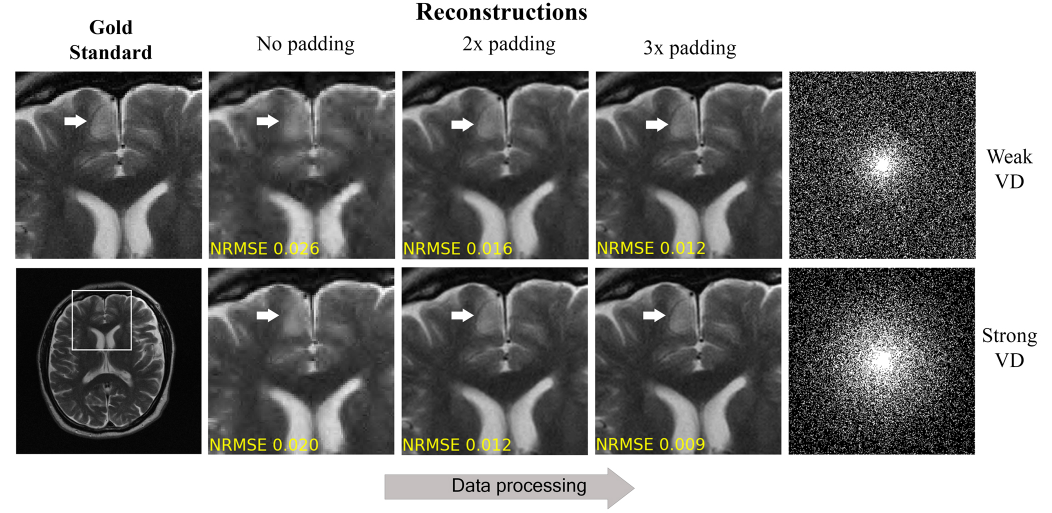}
\caption{Example for \emph{subtle data crime I}: CS reconstructions from retrospectively-subsampled k-space of processed images. Notice how the reconstruction quality improves (both visually in terms of NRMSE) with the zero-padding (data processing) extent. This improvement is completely artificial; it stems from the coupling of early processing and retrospective subsampling which leads to an increased sampling of "true" non-padded data (as illustrated in Figure \ref{fig:fig2}). The artificial improvement is more significant when the sampling is stronger around k-space center (bottom row, strong VD).}. 
\label{fig:fig3}
\end{figure*}

\section*{Subtle data crimes}

In this section we lay out the framework for our experiments.

\subsection*{Subtle Crime I: zero-padded k-space data}
We first consider a data processing pipeline  that is implemented inside many commercial MRI scanners to reconstruct the scanner output (i.e. the MR image). The k-space data are typically acquired using a multi-coil array and the pipeline includes the following steps  (Figure \ref{fig:fig1}a): (1) image interpolation, implemented by zero-padding the raw multi-coil k-space data; (2) application of the inverse Discrete Fourier Transform (DFT), and (3) multi-coil image combination via a square Root Sum of Squares (RSS) step. Notice that although the acquired data are complex-valued, the RSS step produces a magnitude (real and non-negative) image. The scanner output is therefore an interpolated real-valued non-negative image; this is the type of images most prevalent in online MRI databases. 

Let us assume that the scanner image is later downloaded and used for synthesizing new k-space data, with the aim of using those data for training a reconstruction algorithm. The synthesized k-space has two interesting features not originally present: it is larger than the original raw k-space (due to the zero-padding), and it has non-zero values everywhere (due to the non-linear RSS step). In other words, the "true" data now lie in the k-space center, while artificial data appear in its periphery (Figure \ref{fig:fig1}a). However, since this k-space looks fully sampled, it is considered as "ground truth" and used for algorithm development. 

A research pipeline that is commonly used in the development of MRI reconstruction algorithms is based on retrospective-subsampling, where sub-Nyquist sampling is simulated using a binary sampling mask and applied to a fully-sampled k-space (Figure \ref{fig:fig1}c). In the scenario of \emph{subtle data crime I}, such retrospective subsampling is applied to the synthesized k-space, which includes artificial data. Common subsampling masks are typically based on Variable-Density (VD) sampling schemes, which sample the center of k-space more densely than its periphery; VD schemes are used because they produce incoherent aliasing artifacts that can be removed by sparsity-promoting optimization-based reconstruction algorithms \cite{lustig2007sparse}. Importantly, because k-space was zero-padded earlier in the pipeline, application of a VD mask to the entire area of the synthesized k-space results in higher effective sampling density of the "true" k-space data. 

To demonstrate this, we performed the following experiment: we generated subsampling masks for combinations of three zero-padding factors and three sub-sampling schemes (Figure \ref{fig:fig2}a). All the masks included a global sampling rate of $17\%$, which corresponds to an acceleration factor of R=6; this rate is measured for the full k-space area. Then, we measured the \emph{effective} sampling rate, which we define as the sampling rate in the non-padded areas only (yellow boxes in Figure \ref{fig:fig2}a), and plotted it against the zero-padding rate (Figure \ref{fig:fig2}b). The results indicate that for VD subsampling (both weak-VD and strong-VD), the effective sampling rate is much higher than the global rate. In the case of 2x zero-padding, which is often applied \emph{by default} in commercial scanners, the effective rates were 24\% (R=4.1) and 38\% (R=2.6) for weak and strong VD sampling respectively, i.e. much denser than the global rate of 17\% (R=6). Nevertheless, since this subtle effect is often missed by researchers, only the global rate is reported, and it is claimed that algorithms are suitable for reconstruction for a sub-sampling rate that is much larger than the one that was used in practice.  

In summary, our experiment demonstrates that when processed data are retrospectively subsampled with a VD scheme, there is increased sampling density of "true" data. In the experiments described in the next section we demonstrate that this gives rise to overly-optimistic algorithm performance. 

\subsection*{Subtle Crime II: JPEG-compressed data}
The second studied pipeline involves JPEG compression of the scanner image (Figure \ref{fig:fig1}b). Such compression is commonly used to reduce storage footprint, and it is sometimes applied as part of the DICOM data saving pipeline, which is highly prevalent for storage of medical images. To demonstrate the JPEG effect, here we neglect the zero-padding scenario, although the two effects are sometimes combined. In the scenario of \emph{subtle data crime II}, the JPEG-compressed image is stored in an online database and later downloaded and used for synthesizing a new k-space, which is used for algorithm development (Figure \ref{fig:fig1}c). However, since JPEG compression reduces the data entropy, using JPEG data in retrospective-subsampling experiments leads to improved reconstruction fidelity; we aim to show that this leads to an artificial improvement of image reconstruction algorithms.

\section*{Results}
We studied the effects of the hidden data processing pipelines by simulating those pipelines and conducting a large-scale study using the carefully-controlled processed data. Implementation details are provided in the Materials and Methods section. 

\begin{figure*}[!ht]
\centering
\includegraphics[width=1\textwidth]{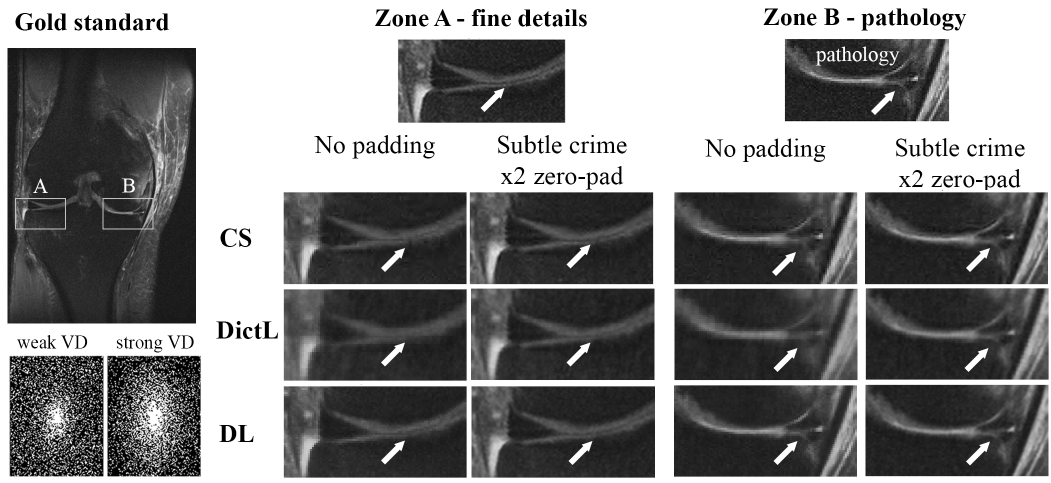}
\caption{\emph{Subtle data crime I}. The CS, DictL and DL algorithms were trained and tested using two versions of the same knee MRI dataset, processed without zero-padding and with 2x zero padding. In the latter case, which represents the scenario of \emph{subtle data crime I}, the reconstructions exhibit sharper images, with improved visibility of small clinically-relevant details. This illustrates that training inverse problem solvers using processed data may lead to overly-optimistic results.}
\label{fig:fig4}
\end{figure*}

\subsection*{\emph{Subtle crime I}}

The first experiment examines the effect of the commercial-scanner data processing pipeline (Figure \ref{fig:fig1}a) on the CS algorithm. The results show that this algorithm produces increasingly sharper reconstructions as the k-space zero-padding factor grows, for both weak and strong VD sampling schemes (Figure \ref{fig:fig3}). This effect is reflected by an artificial reduction of the NRMSE as a function of the zero-padding. 

In the second experiment we implemented the three algorithms and applied them to two versions of the same knee MRI dataset: one prepared without zero-padding, and the other prepared with 2x zero-padding. The algorithms were trained on each dataset separately, and then tested with the corresponding version of a test image that includes fine details and a knee pathology (Figure \ref{fig:fig4}). As can be seen, all the algorithms produced sharper images in the \emph{subtle data crime I} scenario, where the data were zero-padded: the fine details and the pathology became more visible than in the non-padded case. 

\begin{figure*}[!ht]
\centering
\includegraphics[width=1.05\textwidth]{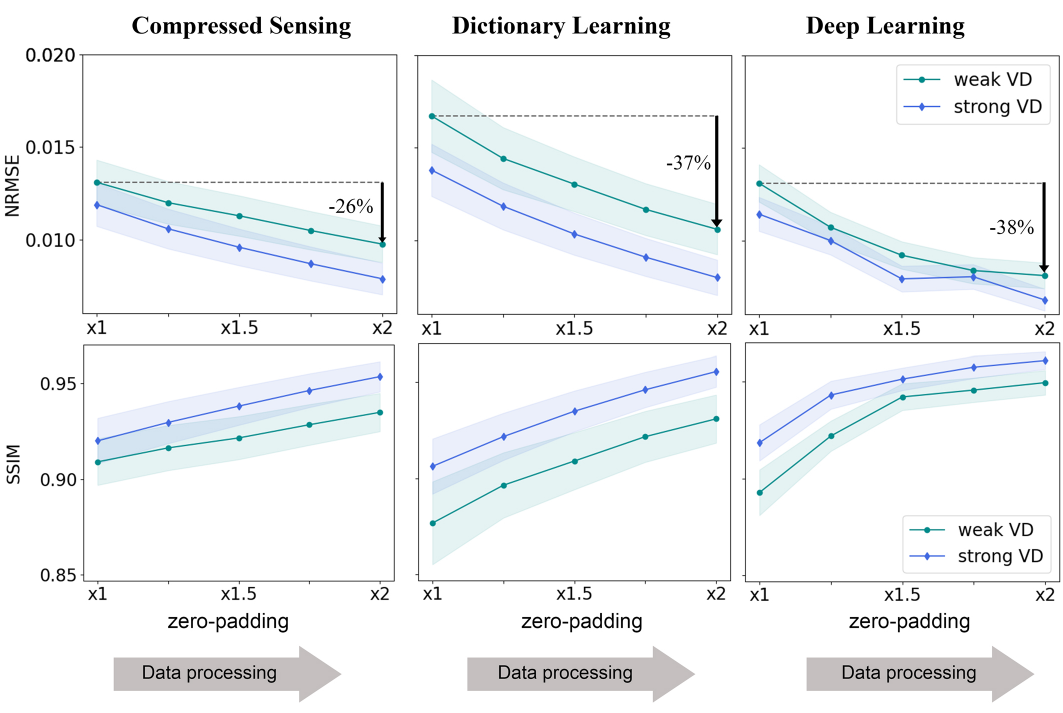}
\caption{\emph{Subtle data crime I} statistics. The CS, DictL and DL algorithms were trained and evaluated using data with various data processing extents. The processing pipeline, which is typically implemented inside commercial scanners, includes k-space zero-padding (Figure \ref{fig:fig1}a). Retrospective subsampling experiments were performed with Variable Density (VD) subsampling with $R=4$. The curves display the mean and STD of the NRMSE and SSIM error metrics for the test set. Notice that both metrics show an artificial improvement that is correlated with the data processing extent. This demonstrates that algorithms evaluated on retrospectively-subsampled processed data tend to yield overly-optimistic evaluation.}
\label{fig:fig5}
\end{figure*}

These results were further confirmed in a large set of experiments, where the algorithms were trained and tested on five versions of the underlying knee dataset representing five data processing scenarios; each dataset contained 2971 images. The hyperparameter calibration, training and testing was performed for each dataset separately, to optimize the algorithmic results for each data processing scenario. We then computed the statistics of two image quality metrics, the NRMSE and Structural Similarity Index (SSIM) \cite{wang2004image}, and plotted them against the zero-padding rate. Markedly, the results of the three algorithms exhibit the same behaviour: their NRMSE and SSIM values improve consistently with the zero-padding extent (Figure \ref{fig:fig5}). This improvement is completely artificial and stems only from the data processing. Strikingly, for the 2x zero-padding case, which is often the default in commercial scanners, the NRMSE exhibits a large improvement of 26$\%$-42$\%$ (Table \ref{tab1:tab1}).

\subsection*{\emph{Subtle crime II}}
To demonstrate the JPEG compression effect, we performed experiments in which the algorithms were trained and tested on different versions of the same underlying dataset. The JPEG compression level is determined by a Quality Factor (QF), where $QF=75$ is the default (that yields lossy compression), and values such as $QF=50$ and $QF=20$ yield increasingly lossy compression \cite{jpeg}. For reference, our experiments also include the case of image reconstruction from Non-Compressed (NC) data. In all cases, the hyperparameter calibration, algorithm training and inference were done on the same type of data (i.e. with NC or a specific QF).

\begin{figure*}[ht]
\centering
\includegraphics[width=1.05\textwidth]{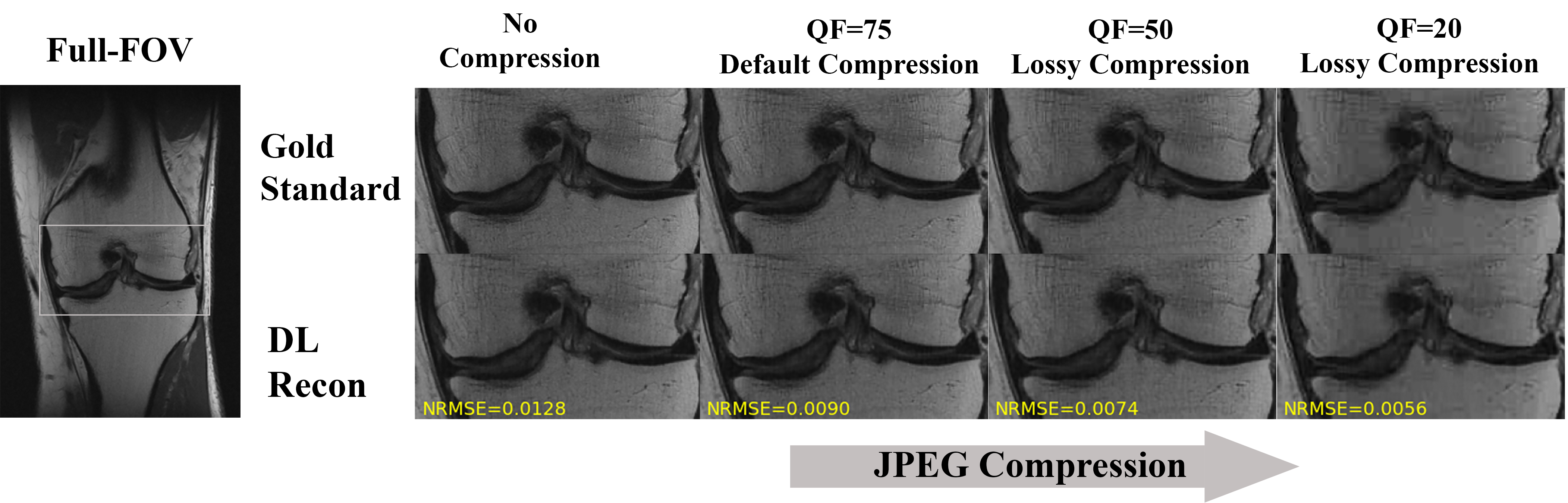}
\caption{Example for \emph{subtle data crime II}. The DL algorithm was trained and tested on non-compressed and JPEG-compressed data. Although the compression reduces the visual image quality, the NRMSE surprisingly reduces with increased compression, reflecting a seemingly-better image quality. The reason is that in retrospective experiments both the "gold standard" and reconstructed images are based on processed data, hence the error metric is blind to the data processing and prone to bias. Strikingly, although the reconstructions from non-compressed and default-compressed data are visually similar, the NRMSE of the latter is lower by 30\%. This demonstrates the subtle bias induced by training and evaluating algorithms on JPEG-compressed data.}
\label{fig:fig6}
\end{figure*}

In the first experiment, the DL algorithm was trained on the different datasets. Figure \ref{fig:fig6} displays an example from the test set, which shows the gold standard images and the DL reconstructions for data undersampled with $R=4$. Generally the visual quality of all the images (both gold standard and reconstructed ones) reduces with increased JPEG compression level (left-to-right in Figure \ref{fig:fig6}); this is expected from compressed data. However, the NRMSE metric shows an unexpected effect: it improves with the compression, i.e. the reconstruction error reduces although the image visual quality degrades. The reason for this phenomenon is that in retrospective experiments the reconstruction quality is measured w.r.t. to a "gold standard" image that is also processed (see the pipeline in Figure \ref{fig:fig1}c); the error metric is therefore blind to data processing. Strikingly, the NRMSE could show a misleading improvement even when the human eye cannot see any difference, as demonstrated in the left two columns of  Figure \ref{fig:fig6}: although the reconstructions from NC and $QF=75$ are visually similar, the NRMSE of the latter is lower by 30$\%$. This reflects the subtle bias induced by the pipeline of \emph{subtle data crime II}.  

\begin{figure*}[ht]
\centering
\includegraphics[width=1.05\textwidth]{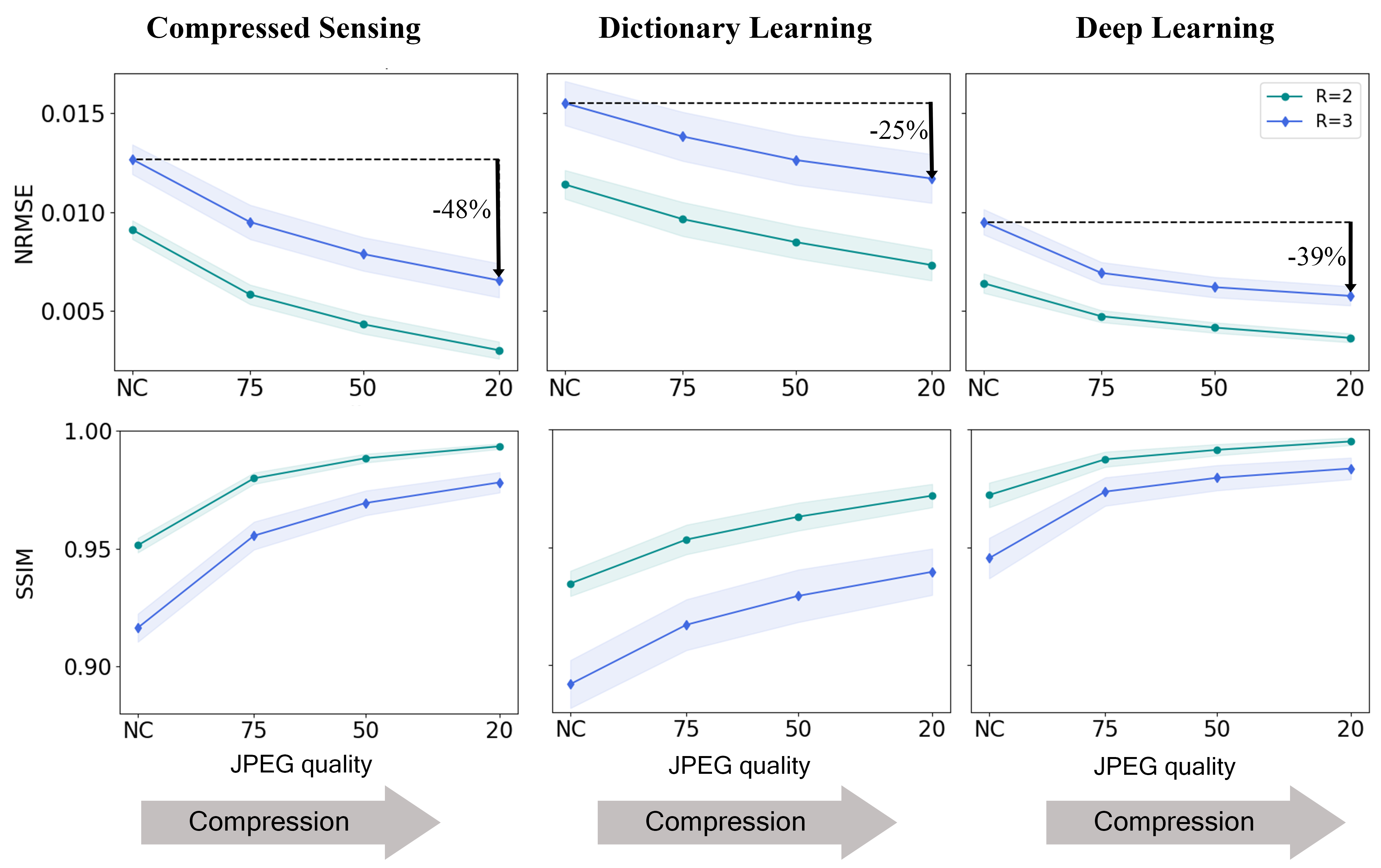}
\caption{Results of a large-scale experiment demonstrating \emph{subtle data crime II}. The CS, DictL and DL algorithms were applied to datasets with No Compression (NC) and increasing JPEG compression levels. The graphs depict the mean and STD computed for the test set. Notice that all the curves show the same trend: the error metrics improve consistently with increased JPEG compression. This improvement is artificial and  stems only from the data processing, which reduces the data entropy. The results therefore demonstrate the subtle bias caused by training inverse problem solvers on JPEG-compressed data.
}
\label{fig:fig7}
\end{figure*}

The JPEG compression effect was further observed in a statistical analysis of a large-scale experiment, where the algorithms were trained and tested on the four types of data (NC, QF=75, QF=50, QF=20) (Figure \ref{fig:fig7}). As illustrated, the error metrics exhibit a consistent improvement with the compression; notably, this effect is systematically observed for all the studied algorithms and reduction factors (Table \ref{tab:tab2}).

\begin{table*}[]
    \centering
    \begin{tabular}{ |p{3cm}|p{1.6cm}|p{1.6cm}|p{1.6cm}|p{1.6cm}|p{1.6cm}|p{1.6cm}|  }
     \hline
     \multicolumn{1}{|c|}{} &\multicolumn{3}{|c|}{NRMSE} &\multicolumn{3}{|c|}{SSIM} \\
     \hline
      & No Padding & 2x padding \ Subtle crime & Artificial \ change & No Padding & 2x padding \ Subtle crime & Artificial \ change \\
     \hline
     CS - weak VD & 0.0131    & 0.0098 & -26$\%$ & 0.91 & 0.93  & 3$\%$  \\
     CS - strong VD & 0.0119    & 0.0079 & -34$\%$ & 0.92 & 0.95  & 4$\%$ \\
     DictL - weak VD &  0.0167 & 0.0106 & -37$\%$ & 0.88 & 0.93  & 6$\%$ \\
     DictL - strong VD &  0.0138 & 0.008 & -42$\%$ & 0.91 & 0.96  & 5$\%$ \\
     DL - weak VD & 0.0131 & 0.0081 & -38$\%$ & 0.89 & 0.95  & 6$\%$\\
     DL - strong VD & 0.0114 & 0.0068 & -41$\%$ & 0.92 & 0.96  & 5$\%$\\
     \hline
    \end{tabular}
    \caption{\emph{Subtle data crime I} statistical results: the mean NRMSE and SSIM values measured for the test set. All three algorithms yield overly-optimistic results when trained and evaluated on zero-padded (processed) MRI data.}
    \label{tab1:tab1}
\end{table*}

\section*{Discussion}

This study reveals that naive usage of open-access data in development of MRI reconstruction algorithms could give rise to overly-optimistic results. The underlying cause is that open-access data are commonly prepared with hidden data processing pipelines that implicitly affect the data properties. Our large-scale study demonstrates that CS, DictL and DL algorithms exhibit biased results for data prepared with common data processing pipelines. Since this form of bias is largely unknown, it is frequently not addressed in research literature; we introduce a framework for studying such bias and coin the term \emph{subtle data crimes} to facilitate research in this field. 

This work offers insights into subtle mechanisms that lead to biased performance of modern reconstruction algorithms. Our main observation is that such bias stems from the unintentional coupling of hidden data processing pipelines with later retrospective-subsampling experiments; the data processing implicitly improves the inverse problem conditioning, and the retrospective subsampling enables the algorithms to benefit from that. This process may appear in different forms. In \emph{subtle data crime I}, the zero-padding concentrates the "true" k-space data to the center, and when VD sampling is later applied, those data are densely sampled; the increased amount of "true" data that becomes available to the algorithm makes the inverse problem easier to solve, hence algorithms tend to exhibit misleadingly-good results. In \emph{subtle data crime II}, the JPEG compression reduces the data entropy, i.e. it increases their sparsity and yields a more compact representation in a sparsifying transform domain. Modern reconstruction algorithms leverage sparsity priors or learn the compact representation from training data \cite{lustig2007sparse,elad2010sparse,fessler2020optimization,zhu2018image}; therefore, they benefit from the compression and yield biased results.  

Another main insight from this study is that in retrospective-subsampling experiments, the error metrics might show a misleading evaluation. That occurs because they measure the difference between two images (the gold standard and reconstructed image) that are based on the same processed data. Ideally, the error metrics should measure the difference between the reconstructed image and the original unprocessed one, but because the latter is unavailable (since it was not stored in the database), the  metrics become blind to the data processing. As a result, they cannot reflect the true reconstruction quality, and they might produce misleading results.

This study also sheds light on a new type of sensitivity of MRI reconstruction algorithms. At present there is growing interest in identifying sensitivities of such algorithms  \cite{antun2020instabilities,cheng2020addressing,darestani2021measuring,genzel2020solving,raj2020improving,koonjoo2021boosting}. However,
recent studies focused mainly on investigating sensitivities with respect to adversarial attacks. While these attacks are an important research tool, they are not observed in practice since MRI scanners are closed systems. Here, on the other hand, we focused on sensitivity related to a more common cause: off-label usage of public databases. While reviewing papers, we noticed that such usage is becoming increasingly more common due to the growing availability of public databases that offer various types of MRI data. \emph{Subtle data crime I} may be common since MR images found in public databases are often based on images produced by commercial scanners, where the data processing pipeline described in Figure \ref{fig:fig1}a is often applied by default. Additionally, \emph{subtle data crime II} may be common since JPEG images are highly prevalent; 73.3\% of the Internet websites contain JPEG-format data \cite{jpeg_usage}. These factors suggest that the \emph{subtle data crimes} might be more common that intuitively expected.

It is worth mentioning that this work did not aim to benchmark the studied algorithms; instead, it aimed to show they are all affected similarly by the \emph{subtle data crimes}. However, as a side benefit, we did obtain benchmark comparisons. To ensure a fair comparison, we dedicated significant efforts to calibrating the hyperparameters of each algorithm for each processed version of the underlying dataset separately (see Materials and Methods); specifically, we dedicated one month of computations to tuning the DictL algorithm parameters through a vast search over a huge search space. Moreover, we ensured that the algorithms were calibrated, trained and tested using identical datasets. 
We empirically observed that the studied algorithms perform overall on-par, with an advantage of CS over DictL and a slight advantage of DL over both. However, due to the pipelines of the \emph{subtle data crimes}, all of our computations were performed with single-coil magnitude non-negative images; the benchmarking of the algorithms for multi-coil, complex-valued MRI data is beyond the scope of this work and remains for future research.

In summary, this research aims to raise a red flag regarding naive off-label usage of open-access data in development of machine learning algorithms. We showed that such usage may lead to biased results of inverse problem solvers. Furthermore, we demonstrated that training MRI reconstruction algorithms using such data could yield an overly-optimistic evaluation of their ability to reconstruct  small clinically-relevant details and pathology; this increases the risk of translation of biased algorithms into clinical practice. We therefore call for attention of researchers and reviewers; data usage and pipeline adequacy should be considered carefully, reproducible research should be encouraged, and research transparency should be required. By introducing the framework for studying \emph{subtle data crimes} we hope to raise community awareness, stimulate discussions and set the ground for future studies of data usage.

\begin{table*}[]
    \centering
    \begin{tabular}{ |p{2cm}|p{1.5cm}|p{1.5cm}|p{1.5cm}|p{1.5cm}|p{1.5cm}|p{1.5cm}|  }
     \hline
     \multicolumn{1}{|c|}{}
     &\multicolumn{3}{|c|}{NRMSE} &\multicolumn{3}{|c|}{SSIM} \\
     \hline
      & No JPEG Compression  & JPEG QF 20 \ Subtle Crime & Artificial \ change &  No JPEG Compression & JPEG QF 20 \ Subtle Crime & Artificial \ change \\
     \hline
     CS - R=2 & 0.0091 & 0.0030 & -67$\%$ 
     & 0.95 & 0.99  & +4$\%$  \\
     CS - R=3 & 0.0127 & 0.0065 & -48$\%$ 
     & 0.92 & 0.98  & +7$\%$  \\
     CS - R=4 & 0.0149 & 0.0091 & -39$\%$ 
     & 0.89 & 0.96  & +8$\%$  \\
     DictL - R=2 & 0.0114 & 0.0073 & -36$\%$ &
     0.93 & 0.97  & +4$\%$  \\
     DictL - R=3 & 0.0155 & 0.0117 & -25$\%$ & 
     0.89 & 0.94 & +5$\%$  \\
     DictL - R=4 & 0.0195 & 0.0166 & -15$\%$ & 
     0.85 & 0.90 & +6$\%$  \\
     DL - R=2 & 0.0064    & 0.0036 & -44$\%$ & 
     0.97 & 0.99  & +2$\%$  \\
     DL - R=3 & 0.0095    & 0.0058 & -39$\%$ & 
     0.95 & 0.98  & +4$\%$  \\
     DL - R=4 & 0.0111 & 0.0071 & -36$\%$ & 
     0.93 & 0.97  & +5$\%$  \\
    
     \hline
    \end{tabular}
    \caption{\emph{Subtle data crime II} statistics: the mean NRMSE and SSIM values measured for the test set. All three algorithms yield overly-optimistic results when trained and evaluated using JPEG-compressed data.}
    
    \label{tab:tab2}
\end{table*}

\section{Materials and Methods}

\subsection*{Raw Data}

To demonstrate the effects of the hidden data processing pipelines, we took raw MRI data and "spoiled" them with carefully-controlled processing steps. The raw data were obtained from the FastMRI database \cite{knoll2020fastmri}. This section describes the raw datasets; the data processing steps were described in the main part of the paper for each \emph{subtle data crime} separately.

\emph{1. Brain data}. In the experiment presented in Figure \ref{fig:fig3} we used a single $320\times320$ brain image.

\emph{2. Knee Fat-Saturated Proton Density (FSPD) data}.
In the knee pathology experiment (Figure \ref{fig:fig4}) we used data from multi-coil FSPD scans, since knee pathology is usually observed in this type of MRI scans. The training set consisted of 2849 randomly-chosen slices obtained from 300 subjects, and the test case was a specific image that contains a pathology (shown in Figure \ref{fig:fig4}). 

\emph{3. Knee Proton Density (PD) data}. In the large-scale experiments that were done for demonstrating the two \emph{subtle data crimes} (Figures \ref{fig:fig5}-\ref{fig:fig7}), we used data from multi-coil Proton Density (PD) scans. Specifically; we used 1427 slices obtained from 80 subjects for training and 122 slices obtained from 7 subjects as the test set. All the slices were chosen randomly.

When constructing the knee PD and FSPD datasets we used only slices from central anatomical regions, i.e. edge slices that contain mostly noise were removed. Additionally, for each dataset, we chose 10 random slices obtained from 10 different subjects and reserved them for tuning the hyperparameters of the studied algorithms; these slices were not included in the training or test sets. It is worth mentioning that the limited number of slices used for hyperparameter calibration was dictated by the need to perform vast computations over a huge search space, especially for the DictL algorithm, as described below.

\subsection*{Experimental overview}
We designed our research framework such that it would enable isolating the bias related to the \emph{subtle data crimes} in a controlled setup. Additionally, since a side-result of this study was the benchmarking of the studied algorithms, we also dedicated significant efforts to ensuring their fair comparison. Here we detail the steps that were taken for these two aims. 

First, to mimic a scenario in which users download a database from an online resource and then optimize the parameters of their algorithm for that specific dataset, we prepared \emph{separate} processed datasets for each instance of the data processing parameters (i.e. for each zero-padding factor or JPEG QF), and ensured that there is no mixture between the datasets. We then calibrated, trained and tested the algorithms on each processed dataset separately. This ensured that each algorithm was evaluated using  instance-optimal parameters; it therefore mitigated bias related to hyperparameter tuning. Secondly, we applied the three algorithms to identical datasets; their results are therefore comparable. Finally, we generated sampling masks on-the-fly, i.e. a different random mask was generated for each k-space example during the training and test sessions. This technique enables generating a large number of sampling masks while maintaining their statistics, hence it prevents over-fitting to any particular sampling mask.

\subsection*{Sampling}
In the retrospective-subsampling experiments, we generated random 2D subsampling masks from pre-defined PDFs using Monte-Carlo experiments. We implemented three subsampling schemes: (1) random-uniform, in which the PDF was constant and equal to $1/R$ ($R$ is the acceleration factor); (2) \emph{weak VD}, in which the PDF was constructed by the function $f(r)=(1-r)^p$, where $r$ is the distance from k-space center and $p$ is the power \cite{lustig2007sparse}, which was set to $p=7$ in this case; and (3) \emph{strong VD}, in which the PDF was also constructed by $f(r)=(1-r)^p$ and the power was set to $p=1$, $p=2$ and $p=3$ for reduction factors of $R=2$, $R=3$, and $R=4$ correspondingly. All the sampling masks included a small fully-sampled area in the center of k-space. In parallel imaging this area is often known as the \emph{calibration region} \cite{griswold2002generalized}, and in single-coil MRI experiments this region ensures sampling of the low-frequency data and helps stabilize the computational results. The calibration region size was $12\times7$ pixels for the $640\times372$ knee images and $6\times6$ pixels for the $320\times320$ brain image. In the zero-padding experiments, where the image size varied, the calibration region size scaled with the image size.

\subsection*{Algorithms}
The CS, DictL and DL algorithms recover an MR image from subsampled k-space measurements by solving an inverse problem that has the following general form:

\begin{equation}
    \hat{\mathbf{x}} = \arg\min_{\mathbf x} \frac{1}{2}\left\lVert \mathbf{Ex}-\mathbf{y}\right\rVert^2_2 +\mathbf{\lambda} R(\mathbf{x}),
    \label{eq:eq1}
\end{equation}
where $\mathbf{x}$ is the image to be reconstructed, $\mathbf{y}$ are the k-space measurements, $\mathbf{E}$ is an encoding operator that describes the  imaging system, $\mathbf{R({x})}$ is a regularization term, and $\mathbf{\lambda}$ is trainable parameter that controls the tradeoff between the Data Consistency (DC) term (the first term in Eq. [\ref{eq:eq1}]) and the regularization term. In MRI, the encoding operator $\mathbf{E}$ is typically described as $\mathbf{E = UF}$, where $\mathbf{F}$ is the Fourier transform and $\mathbf{U}$ is an operator that describes the k-space subsampling. The studied algorithms differ in their regularization terms and optimization techniques, as described next.

\emph{\textbf{CS algorithm}}. This algorithm formulates eq. [\ref{eq:eq1}] as a convex optimization problem with an $\ell_1$ prior that promotes the sparsity of $\mathbf{x}$ in a sparsifying transform domain \cite{lustig2007sparse}. A common choice for the prior is an $\ell_1$-wavelet one; the optimization problem is then,
\begin{equation}
    \hat{\mathbf{x}} = \arg\min_{\mathbf x} \frac{1}{2}\left\lVert \mathbf{Ex}-\mathbf{y}\right\rVert^2_2 +\lambda\Vert\mathbf{\Psi} \mathbf{x}\Vert _1,
    \label{eq:eq2}
\end{equation}
where $\mathbf{\Psi}$ is the wavelet transform. Eq. [\ref{eq:eq2}] can be solved using different optimization techniques; here it was solved using the Fast Iterative Shrinkage-Thresholding Algorithm (FISTA) \cite{beck2009fast}. Our implementation was based on the SigPy python toolbox \cite{ong2019sigpy}. 

The CS algorithm has one tunable parameter, $\mathbf{\lambda}$. We calibrated it through a grid search, where the grid included values in $\mathbf{\lambda} \in [1e-9,1e-1]$. We ran the grid search over 10 images from a subset of the data that was reserved for hyperparameter tuning. We then computed the mean NRMSE over those 10 images, and the value of $\mathbf{\lambda}$ that corresponded to the lowest mean NRMSE was chosen. Since in the experiments of \emph{subtle data crime I} the image size varies with the zero-padding, we repeated this procedure for each image size separately. However, we empirically observed that the same $\mathbf{\lambda}$ value was chosen for all image sizes. The chosen values were $\mathbf{\lambda}=0.005$ for the brain data (Figure \ref{fig:fig3}) and $\mathbf{\lambda}=0.001$ for the knee data (Figures \ref{fig:fig4}-\ref{fig:fig7}).

\emph{\textbf{DictL algorithm}}. The DictL algorithm reconstructs the image $\mathbf{x}$ by jointly learning a dictionary $\mathbf{D}$ and a sparse code $\mathbf{A}$. The dictionary, which is used for reconstruction of patches, is a sparsifying transform that is learned directly from the subsampled k-space data. In this method, the image is reconstructed by representing it as a sparse linear combination of dictionary atoms, $\mathbf{x=DA}$. The DictL algorithm jointly solves for the image $\mathbf{x}$, the dictionary $\mathbf{D}$, and the sparse code $\mathbf{A}$ \cite{ravishankar2010mr}. The algorithm learns the dictionary adaptively from the subsampled k-space while reconstructing the image, i.e. it is trained without any other examples or without access to the fully-sampled k-space data; the learning is done over image patches. 

The optimization problem is formulated as follows:
\begin{equation}
\begin{split}
\min_{\mathbf{x}, \mathbf{A},\mathbf{D}} \frac{1}{2}|| \mathbf{Ex}- \mathbf{y}||_2^2 + \frac{\lambda_D}{2} || \mathbf x - \mathbf{R(DA)}||_2^2\quad
\\
\text{subject to }  
||\mathbf{a}_l || &\leq K, \quad l=1,...,L 
\\
||\mathbf{d}_p ||_2 & \leq 1, \quad p=1,...,P
\end{split}
\label{eq:eq3}
\end{equation}
where $\mathbf{R}$ is a reshaping operator that reshapes patches into an image, $\mathbf{a}_l$ are the columns of $\mathbf{A}$, where each column is a vectorized patch from the image, $K$ is the sparsity level, $L$ is the number of patches that are used as training examples during one iteration of the algorithm, $\mathbf{d}_p$ are the columns of the dictionary $\mathbf{D}$, where each column is a vectorized atom, and $P$ is the number of dictionary atoms. 

We implemented the DictL algorithm in python using our open-source code \cite{tamir2020dict}. The algorithm solves eq. [\ref{eq:eq3}] through alternating minimization, with K-SVD \cite{aharon2006k} for the dictionary update and Orthogonal Matching Pursuit (OMP) \cite{cai2011orthogonal} for the sparse code update. 

We dedicated vast computations for calibrating the five tunable hyperparameters of the DictL algorithm, which are: $P$, $K$, $\lambda_D$, block size $b$ (the blocks are symmetric, i.e. if $b=8$ then the block size is $8\times8$), and the number of outer iterations of the alternating minimization algorithm, $N_{iter}$. Due to the varying image size in the zero padding experiments, we repeated the calibration process for each image size separately; this added another dimension. Furthermore, we repeated the search for a set of five images that was reserved for hyperparameter tuning (the number of images was dictated by the need for vast computations). Altogether, we trained the DictL algorithm five times for each combination of the 6 parameters; this resulted in evaluation of 77,686 DictL instances. For each combination we computed the mean NRMSE over the five images, and chose the hyperparameter values that produced the lowest NRMSE. This set of computations was highly time consuming; it was conducted on 200 CPUs in parallel for over four weeks. The tested grid search values were:
$P\in[100,200,300]$, $K\in[5,13]$ where the step size was 2, $\lambda_D\in[1e-5,1e-4,1e-3,1e-2,1e-2]$, $b\in[4,32]$ where the step size was 4, and $N_{iter}\in[5,13]$ where the steps size was 2. The chosen hyperparameter values are detailed in the Supplementary Material.

\emph{\textbf{DL algorithm}}. We studied the Model-based reconstruction using Deep Learned priors (MoDL)  algorithm, which gives state-of-the-art performance in MRI reconstruction \cite{aggarwal2018modl}. MoDL solves the following optimization problem:
\begin{equation}
    \hat{\mathbf{x}} = \arg\min_{\mathbf x} \frac{1}{2}\left\lVert \mathbf{Ex}-\mathbf{y}\right\rVert^2_2 +\lambda \left\lVert \mathbf{x}-\mathbf{D_w}(\mathbf{x})\right\rVert^2.
    \label{eq:imdl}
\end{equation}
where $\mathbf{D_w}(\mathbf{x})$ is the output of a Convolutional Neural Network (CNN). This optimization problem is solved using an unrolled deep neural network which includes interleaved CNNs and DC blocks. The DC blocks ensure consistency of the solution with the k-space measurements; the backpropagation through them is implemented using the Conjugate Gradient (CG) algorithm \cite{hestenes1952methods}. The MoDL unrolled network is trained in an end-to-end supervised manner, where the input is an aliased image obtained from the zero-filled subsampled k-space data and the target is a "gold standard" image obtained from the fully-sampled k-space. 

In our implementation, the architecture of the network included 6 unrolls, CNNs with a U-Net structure \cite{ronneberger2015u}, weight sharing, and 8 CG steps in the DC blocks. The training was performed using an \emph{l}1-loss and the Adam optimizer \cite{kingma2014adam}, with gradient accumulation such that the effective batch size was 20. The number of epochs was 70. We implemented MoDL using  PyTorch \cite{deepinpy}.

\subsection*{\emph{Subtle data crime I} experiments}
In this section, we provide implementation details regarding the experiments that were performed for demonstrating the effects of \emph{subtle data crime I}. In the first experiment, which demonstrates the difference between global and effective sampling (Figure \ref{fig:fig2}), a set of 15 random masks was generated for each combination of a subsampling scheme and zero-padding factor. The curves in Figure \ref{fig:fig2} depict the mean effective rates measured over those sets. 

In the next experiments, which demonstrate the zero-padding effect (Figures \ref{fig:fig3}-\ref{fig:fig5}), we implemented VD sampling with an acceleration factor of $R=4$. In these experiments, the MoDL network could not be trained on full-size images because the zero-padding enlarges the image size to an extent that poses a computational challenge even with modern GPUs. However, a major advantage of MoDL is that it is convolutional and at inference it can be implemented to any image size \cite{aggarwal2018modl}. We therefore trained MoDL on patches extracted from training images; the patch size was 0.25 of the image size in each dimension, and a single patch was extracted randomly from each image. In contrast, during inference the network was applied to the full-size test images; the results shown in Figure \ref{fig:fig5} therefore represent the reconstruction error for full images. 

\subsection*{\emph{Subtle data crime II} experiments}
In the second set of experiments we studied how the performance of reconstruction algorithms is influenced by JPEG compression of the underlying data. We prepared the processed datasets using the standard JPEG implementation found in the PILLOW library \cite{clark2015pillow}. In the JPEG experiments the reduction factor ranged from $R=2$ to $R=4$ (see Table \ref{tab:tab2}).

\subsection*{Image quality metrics}
We quantified the \emph{subtle data crimes} effects by studying how the data processing pipelines influence two highly common image quality metrics: the Normalized Root Mean Square Error (NRMSE) and the Structural Similarity Index (SSIM) \cite{wang2004image}; the latter was implemented using the SSIM-PIL library \cite{ssim-pil2018}. 

\subsection*{Computational overview}
In this section, we give an overview of the compute times and resources that were used in this research. All of our experiments were performed on 12GB Nvidia Titan Xp GPUs and Intel(R) Xeon(R) Silver 4116 CPUs.

\emph{Hyperparameter tuning}.
The most extensive part of the research was the hyperparameter tuning for the DictL algorithm (described above), which was conducted for over four weeks on 200 CPUs in parallel. The calibration of the DL algorithm hyperparameters was conduced for a similar amount of time using a single GPU. The CS parameter tuning time was only several hours.

\emph{Experiments}. The experiment for demonstrating \emph{subtle data crime I} with a single brain image (Figure \ref{fig:fig3}) required only about several minutes on a standard laptop. In contrast, the experiment for demonstrating this \emph{subtle data crime} using fat-saturated PD knee data (Figure \ref{fig:fig4}) required one day of computations on two GPUs and 40 CPUs. The experiments for obtaining the statistics of \emph{subtle data crime I} (Figure \ref{fig:fig5}) were computationally more demanding, since they required training and testing ten different instances of each algorithm, for the ten combinations of the studied zero-padding ratios and VD sampling schemes. The compute time of these experiments was one week on a single GPU and 200 CPUs. 

Similarly, the experiments that demonstrate \emph{subtle data crime II} (Figures \ref{fig:fig6} and \ref{fig:fig7}) required training and evaluating twelve instances of each algorithm, for the twelve combinations of the four studied compression scenarios and three reduction factors. The CS computation time was several hours. However, the DictL runs were conducted for about eight days on 100 CPUs, and the DL runs were conducted for six days on 12 GPUs.

Altogether, the compute time was about two months using 200 CPUs and 12 GPUs.

\subsection*{Reproducibility}
In the spirit of reproducible research, our code is publicly available here:

\url{https://github.com/mikgroup/subtle_data_crimes}

All the data used in this research is publicly available as part of the FastMRI database \cite{knoll2020fastmri}.

\subsection{Acknowledgment}
The authors acknowledge funding from grants U24 EB029240-01, R01EB009690, R01HL136965. The authors thank Shreyas Vasanawala for his assistance with identifying the pathology cases in the FastMRI data.

\bibliographystyle{ieeetr}
\bibliography{subtle_crimes_paper}  

\end{document}